\def\year{2022}\relax
\documentclass[letterpaper]{article} % DO NOT CHANGE THIS
\usepackage{aaai22}  % DO NOT CHANGE THIS
\usepackage{times}  % DO NOT CHANGE THIS
\usepackage{helvet}  % DO NOT CHANGE THIS
\usepackage{courier}  % DO NOT CHANGE THIS
\usepackage[hyphens]{url}  % DO NOT CHANGE THIS
\usepackage{graphicx} % DO NOT CHANGE THIS
\usepackage{natbib}  % DO NOT CHANGE THIS AND DO NOT ADD ANY OPTIONS TO IT
\usepackage{caption} % DO NOT CHANGE THIS AND DO NOT ADD ANY OPTIONS TO IT
\DeclareCaptionStyle{ruled}{labelfont=normalfont,labelsep=colon,strut=off} % DO NOT CHANGE THIS
\usepackage{algorithm}
\usepackage{algorithmic}
\usepackage{graphicx}
\usepackage{amsmath}
\usepackage{amssymb}
\usepackage{booktabs}
\usepackage{epsfig}
\usepackage{textcomp}
\usepackage{soul}
\usepackage{multirow}
\usepackage{multicol}
\usepackage{newfloat}
\usepackage{listings}
\floatstyle{ruled}
\newfloat{listing}{tb}{lst}{}
\floatname{listing}{Listing}
\title{TANet: A new Paradigm for Global Face Super-resolution via Transformer-CNN Aggregation Network}
\author{
	Yuanzhi Wang\textsuperscript{\rm 1},
	Tao Lu\textsuperscript{\rm 1}\thanks{is the corresponding author},
	Yanduo Zhang\textsuperscript{\rm 1},
	Junjun Jiang\textsuperscript{\rm 2},
	Jiaming Wang\textsuperscript{\rm 3},
	Zhongyuan Wang\textsuperscript{\rm 4},
	Jiayi Ma\textsuperscript{\rm 5}
}
\affiliations{
    \textsuperscript{\rm 1} Hubei Key Laboratory of Intelligent Robot, Wuhan Institute of Technology, Wuhan, China\\
    \textsuperscript{\rm 2} School of Computer Science and Technology, Harbin Institute of Technology, Harbin, China\\
    \textsuperscript{\rm 3} State Key Laboratory for Information Engineering in Surveying, Mapping and Remote Sensing, Wuhan, China\\
    \textsuperscript{\rm 4} School of Computer Science, Wuhan University, Wuhan, China\\
    \textsuperscript{\rm 5} the Electronic Information School, Wuhan University, Wuhan, China
}

\usepackage{bibentry}
\begin{document}

\maketitle

\begin{abstract}
Recently, face super-resolution (FSR) methods either feed whole face image into convolutional neural networks (CNNs) or utilize extra facial priors (e.g., facial parsing maps, facial landmarks) to focus on facial structure, thereby maintaining the consistency of the facial structure while restoring facial details.
However, the limited receptive fields of CNNs and inaccurate facial priors will reduce the naturalness and fidelity of the reconstructed face.
In this paper, we propose a novel paradigm based on the self-attention mechanism (i.e., the core of Transformer) to fully explore the representation capacity of the facial structure feature.
Specifically, we design a Transformer-CNN aggregation network (TANet) consisting of two paths, in which one path uses CNNs responsible for restoring fine-grained facial details while the other utilizes a resource-friendly Transformer to capture global information by exploiting the long-distance visual relation modeling.
By aggregating the features from the above two paths, the consistency of global facial structure and fidelity of local facial detail restoration are strengthened simultaneously.
Experimental results of face reconstruction and recognition verify that the proposed method can significantly outperform the state-of-the-art methods.
\end{abstract}

\begin{figure}[h]
	\centering{\includegraphics[width=\linewidth]{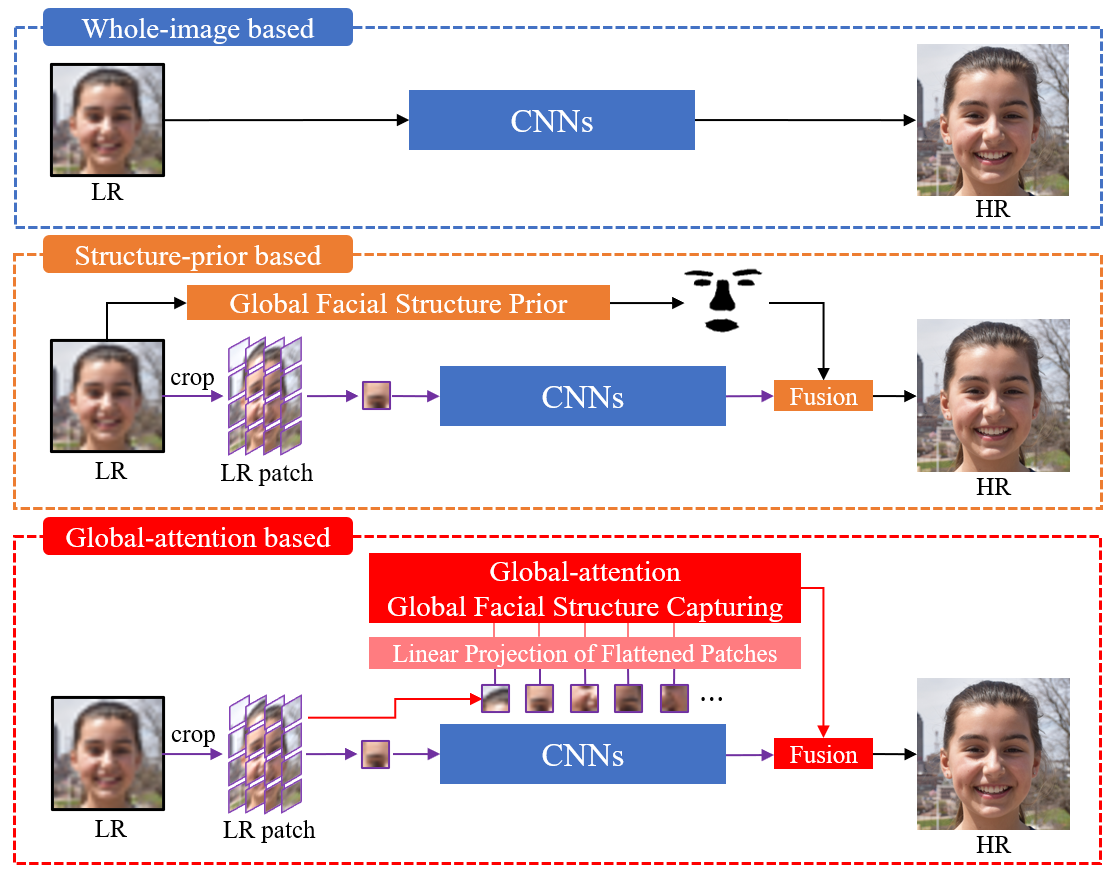}}
	\caption{Three paradigm for global methods. Whole-image based paradigm aims to input the whole face image into CNNs to focus on the global facial structure. Structure-prior based paradigm captures global facial structure by utilizing extra facial structure prior. Global-attention based paradigm leverages the self-attention mechanism and multi-layer perceptron (MLP) to capture the global facial structure.}
	\label{fig:global}
\end{figure}

\section{Introduction}
\label{sec:intro}
%In video surveillance, the far distance between surveillance cameras and face often results in low-resolution (LR) face images, which tremendously affect the face analysis and application. Restoring high-resolution (HR) face images from LR ones by using face super-resolution (FSR) is one of vital technique for intelligent video surveillance system.
Face super-resolution (FSR) is a domain-specific image super-resolution problem, which aims to infer high-resolution (HR) face images from the given low-resolution (LR) ones.
Due to the importance of face in human identity verification, restoring high-quality face images using the FSR technique has attracted much attention in the past decades.

The concept of FSR was originally proposed by \cite{HF}, they designed a multi-level learning and prediction model to infer high-frequency details of face images.
Followed the above work, Liu \emph{et al.} \cite{liutwoSTEP} proposed a two-step FSR method by integrating a global parametric principal component analysis (PCA) model and a local nonparametric Markov random field (MRF) model.
Since then, researchers have proposed various FSR methods based on these two pioneering works.
In the latest survey of FSR \cite{FSRSurvey}, the early classical methods
are divided into three categories: local patch-based methods \cite{localmethod}, global face statistical methods \cite{PCA}, and hybrid methods \cite{twostep}.
The local patch-based methods crop input face images into several LR patches and then restore texture details locally, but they ignore global facial structure information, thus resulting in inconsistent facial contours.
Compared with a general image, the face image is a highly structured object, thus using structured information to guide face reconstruction is an essential paradigm, especially the reliable global facial structure is very important for downstream tasks such as face detection and recognition \cite{MTCNN,mobilefacenets}.
%For this reason, global face statistical methods extract global face structural information on the whole face image to guide face reconstruction.
The hybrid methods combine above two methods to reconstruct both global face contour and local facial details. However, due to their complexity and difficulty of implementation, researchers are mainly focusing on local or global methods

Recently, convolutional neural networks (CNNs) have dominated the FSR task, thanks to their unprecedented local modeling capacity to predict fine-grained facial details.
According to whether global structure information is considered, the CNNs-based FSR methods can be divided into two categories: local methods \cite{regionbasedFSR,PRDRN,SISN} and global methods \cite{BCCNN,FSRNet,DIC}.
The local methods feed the different local regions (blocks) of face image into the CNNs to infer missing facial high-frequency details locally.
%However, the lack of global facial structure information can lead to generated face images with fuzzy facial structure.
However, due to the destruction of the global facial structure information caused by the blocking operation, the generated facial image will inevitably appear blurry effects.
Therefore, some researchers are seeking to develop global methods to focus on and capture the global facial structure, thereby maintaining the global facial structure.
Two popular paradigms for global methods (whole-image based and structure-prior based) are summarized in Figure \ref{fig:global}, where the whole-image based paradigm aims to feed the whole face image into CNNs to focus on the global facial structure \cite{BCCNN,Waveletsrnet}.
However, this paradigm is inherently not well suited for capturing the global information from an input image, because the vanilla CNNs cannot process long-range dependencies (limited receptive fields).
In contrast, the structure-prior based paradigm can better capture global facial structure by utilizing the location of facial components and facial structure information provided by extra facial priors (e.g., facial parsing maps, facial landmarks) \cite{FSRNet,DIC}.
However, the structure-prior based paradigm also suffers from some limitations.
The extra facial priors often require additional prediction models, which inevitably increase the cost and difficulty of training and inference. In addition, it will also causes difficulty for the model to fit and predict unrealistic prior information, and thus resulting in distorted HR face images.

To effectively take advantage of the global facial structure information, we introduce the Transformers that are capable of capturing long-distance visual relations by self-attention mechanism \cite{dsnet}.
In particular, we propose a novel global-attention based paradigm for FSR (Shown in Figure \ref{fig:global}) that can capture global facial structure by exploring and exploiting the long-distance visual relation modeling of self-attention mechanism and multi-layer perceptron (MLP).
According to the proposed paradigm, we design a novel Transformer-CNN aggregation network (TANet) for FSR to retain both global facial structure and local facial details via two parallel paths.
Specifically, we first propose a self-calibrated multi-path fusion network (SMFN) consisting of pure CNNs as a local representation path of TANet to focus on local facial details.
Then, we design a resource-friendly pure Transformer as a global representation path of TANet to capture global facial structure information while reaching low-cost computation by aligning the embedded dimension with the feature dimension of the local representation path (Training requires only two Nvidia RTX 2080Ti GPUs).
The above two individual parallel paths effectively disentangle the local and global representations, which avoids the ambiguity caused by the local representation of CNNs and the global representation of Transformers and helps to maximize their merits.
Finally, the outputs of the above two paths are aggregated by the proposed global-local fusion module, thereby maintaining the consistency of global facial structure and the fidelity of local facial detail restoration.
The contributions of our work can be summarized as follows:

(\romannumeral1) We propose a novel global-attention based paradigm for global FSR to capture global facial structure by exploring and exploiting the long-distance visual relation modeling of self-attention mechanism and MLP.

(\romannumeral2) We design a novel TANet, which can focus on global facial structure and local facial texture details simultaneously by fusing the global representations from Transformers and local feature extraction of CNNs.

(\romannumeral3) We customize a resource-friendly Transformers to effectively capture global facial structure while reducing the considerable computational costs by aligning the embedded dimension with the feature dimension of the local representation path.
Meanwhile, we designed an SMFN to guarantee the robust local representation capability to predict local facial details by iteratively calibrating the representation of local features, thereby further improving the performance of face reconstruction.

\begin{figure*}[h]
	\centering{\includegraphics[width=16cm]{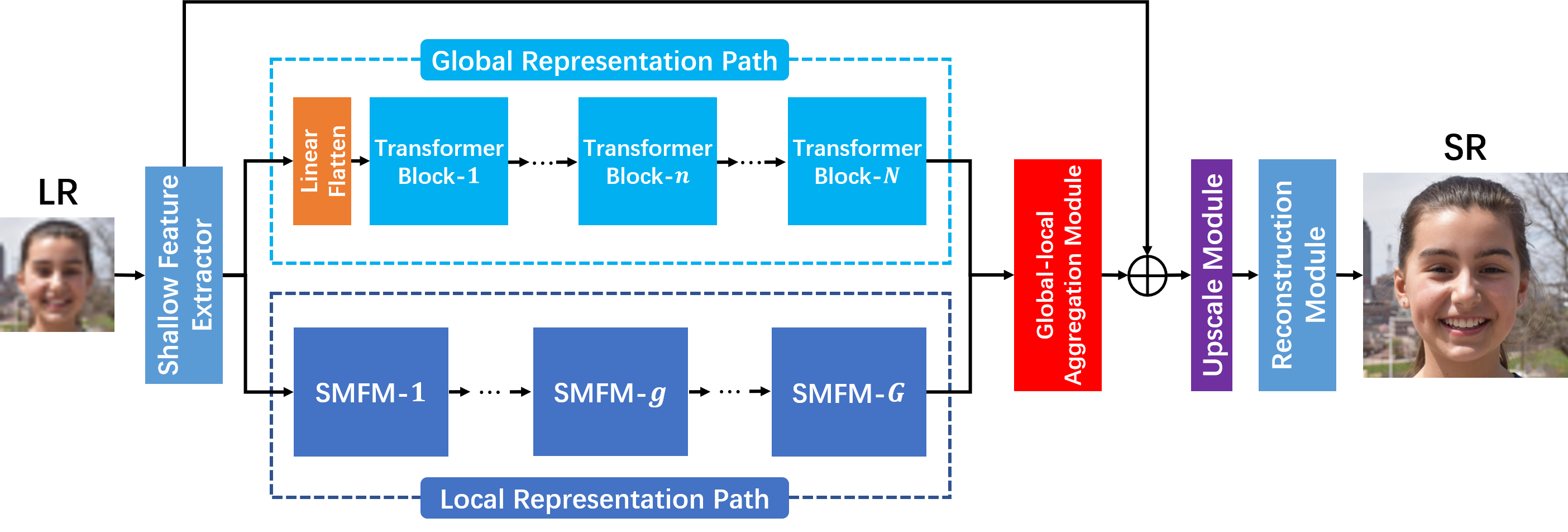}}
	\caption{Overall architecture of TANet, which consists of six components: a shallow feature extractor, a global representation path composed of pure Transformers, a local representation path composed of pure CNNs, a global-local aggregation module, a upscale module, and a reconstruction module.}
	\label{fig:TANet}
\end{figure*}

%-------------------------------------------------------------------------
\section{Related Work}
Transformers originally stemmed from the field of natural language processing (NLP) \cite{transformer}, and its multi-head self-attention and feed-forward MLP layer are stacked to capture the long-range correlation between words.
Hence, the Transformers have become the dominant paradigm in NLP tasks \cite{bert,Transformernlp2,Transformernlp3}.
Motivated by the great success of Transformers in NLP, there are many attempts to explore and exploit the benefits of Transformer in various vision tasks to emphasize the significance of extracting global features.
For example, Dosovitskiy \emph{et al.} \cite{VIT} proposed the Vision Transformer, which views 16$\times$16 image patches as token sequence and predicts category of image via a unique class token.
Jiang \emph{et al.} \cite{transgan} built a Transformer-based generative adversarial network without any convolution operators.

Despite the superiority of aforementioned pure Transformers upon extracting global representations, image-level self-attention is unable to capture local fine-grained details.
To tackle this issue, some existing methods introduce CNNs into vision transformers in different manners because CNNs have unique advantages of local modeling and translation invariance.
For instance, Carion \emph{et al.} \cite{detr} cascaded CNNs and Transformers to conduct end-to-end object detection task.
Chen \emph{et al.} \cite{IPT} presented the Image Processing Transformer for low-level vision task such as image denoising, super-resolution, and deraining, where the CNNs are used to extract shallow-level features and reconstruct images.
However, if the current vision Transformers are directly applied to FSR tasks will suffer from the following two problems: (1) Both pure Transformer and hybrid methods are dominated by the Transformers, which inevitably leads to huge resource consumption. For example, Jiang \emph{et al.} \cite{transgan} proposed a TransGAN for image generation that requires 16 Nvidia Tesla V100 GPUs for training; (2) The primary goal for FSR is to predict lost local texture details.
Compared with Transformers, CNNs have the unique advantage of local modeling to capture fine-grained details.
Therefore, it is extremely inefficient to use Transformers-dominated architectures to implement the FSR task.
In this paper, we customize a CNN-dominated hybrid architecture for FSR to reliably restore local facial detail and capture global facial structure while avoiding huge computational costs.

\section{Transformer-CNN Aggregation Network for Face Super-Resolution}

In this section, we elaborate on the overall architecture of the proposed TANet and the two paths responsible for the global representation and the local representation.

\begin{figure*}[t]
	\centering{\includegraphics[width=15cm]{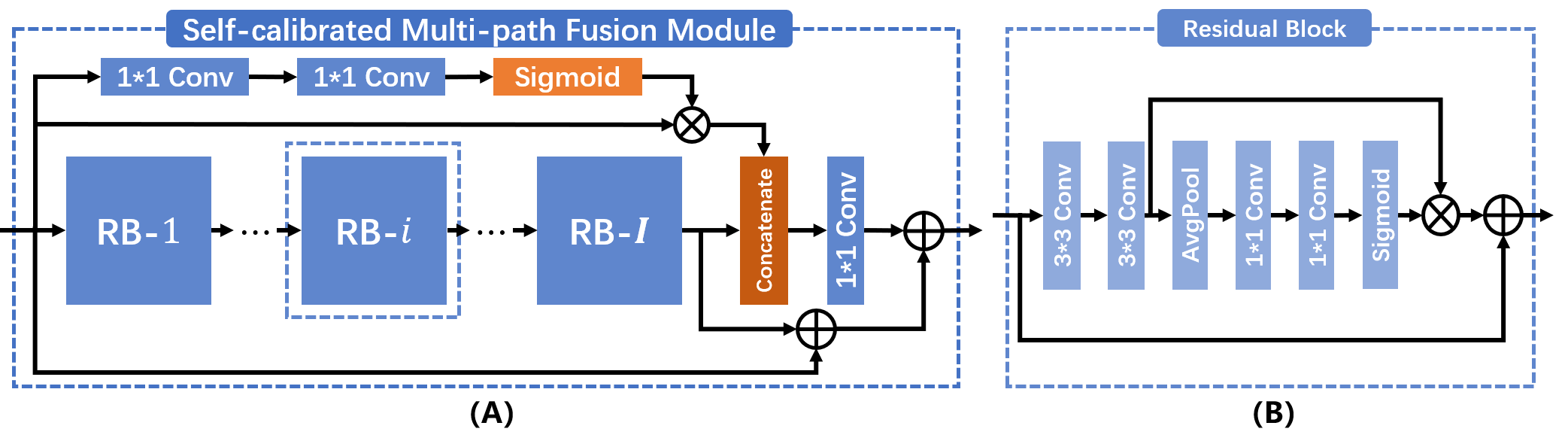}}
	\caption{Architecture of SMFM. (A) shows the overall structure of SMFM, which contains $I$ RBs, a feature calibration path, and the multi-path feature fusion part. (B) shows the detailed structure of RB.}
	\label{fig:SMFM}
\end{figure*}

\subsection{Overall Architecture}
The overall architecture of our proposed TANet is shown in Figure \ref{fig:TANet}, which consists of six components: a shallow feature extractor, a resource-friendly Transformers is established for global representation path, a self-calibrated multi-path fusion network is established for local representation path, a global-local aggregation module, a upscale module, and a reconstruction module.

Let’s denote $I_{LR}$, $I_{SR}$ as the input and output of TANet, and we first use a shallow feature extractor consisting of a 3$\times$3 convolutional layer to extract shallow feature $F_{shallow}$ containing rich facial structure information from the input images.
\begin{equation}
F_{shallow} = H_{shallow}(I_{LR}),
\end{equation}
where $H_{shallow}(.)$ denotes the shallow feature extractor with one convolutional layer.
$F_{shallow}$ is then used as input for the global representation path and local representation path to extract global facial structure and local facial detail, respectively.
So we can further have
%\begin{small}
\begin{equation}
F_{global}\!=\!H_{global}(\!F_{shallow}\!),~F_{local}\!=\! H_{local}(\!F_{shallow}\!),
\end{equation}
%\end{small}
where $H_{global}(.)$ and $H_{local}(.)$ indicate the function of global representation path and local representation path, respectively.
$F_{global}$ and $F_{local}$ denote the global feature extracted by the global representation path and the local feature extracted by the local representation path, respectively.
After obtaining global and local features, we propose a global-local aggregation module (GLAM) to fuse $F_{global}$ and $F_{local}$.
Specifically, the proposed GLAM first reshapes $F_{global}$ into the tensor size of the input feature (i.e., the feature size of $F_{shallow}$) and appends a 1$\times$1 convolutional layer to fine-tune the feature representation.
\begin{equation}
F_{global}^{ft} = H_{Conv}^{1\times1}(F_{global}),
\end{equation}
where $H_{Conv}^{1\times1}(.)$ denotes a convolutional layer with a kernel size of 1$\times$1, $F_{global}^{ft}$ indicates the reshaped and fine-tuned global feature.
Then, we use a 3$\times$3 convolutional layer to fine-tune the local features and output fine-tuned local feature $F_{local}^{ft}$.
\begin{equation}
F_{local}^{ft} = H_{Conv}^{3\times3}(F_{local}),
\end{equation}
where $H_{Conv}^{3\times3}(.)$ denotes a convolutional layer with a kernel size of 3$\times$3.
Finally, we concatenate the global and local features along the channel dimension and simultaneously learn the global and local representations using one 1$\times$1 convolutional layer to output the fused feature $F_{gl}$.
\begin{equation}
F_{gl}= H_{Conv}^{1\times1}(H_{Cat}(F_{global}^{ft}, F_{local}^{ft})),
\end{equation}
where $H_{Cat}(.)$ denotes the concatenating feature operation along the channel dimension.
This function can also be formulated as
\begin{equation}
F_{gl} =  H_{GLAM}(F_{global}, F_{local}),
\end{equation}
where $H_{GLAM}(.)$ denotes the function of GLAM.

After obtaining $F_{gl}$, the upscaling operation needs to be performed by the upscale module.
\begin{equation}
F_{up} = H_{up}(H_{Sum}(F_{gl}, F_{shallow})),
\end{equation}
where $F_{up}$ and $H_{up}(.)$ denote the upscaled feature and a upscale module, respectively.
$H_{Sum}(.)$ denotes an element-wise summation operation.
Finally, the upscaled feature is reconstructed via a reconstruction module and outputs the target HR face image $I_{SR}$. $I_{SR}$ is formulated as:
\begin{equation}
I_{SR} = H_{Recon}(F_{up}) = H_{TANet}(I_{LR}),
\end{equation}
where $H_{Recon}(.)$ and $H_{TANet}(.)$ indicate the reconstruction module
composed of a convolutional layer with a kernel size of 3$\times$3 and the function of our TANet, respectively.
The whole network is optimized with the $L_{1}$ loss function (mean absolute error) which is referenced from \cite{RCAN,SISN}.

\subsection{Resource-friendly Transformers for Global Representation}
Enlightened by \cite{VIT}, the proposed Transformers contains $N$ repeated Transformer blocks, each Transformer block consists of a multi-head self-attention module and a MLP block.
The global representation path in Figure \ref{fig:TANet} shows the structure of the proposed resource-friendly Transformers.
Since the Transformers take embedding token words as inputs and calculate the correlation between each token recursively, the shallow feature $F_{shallow}$ is tokenized into several patches as the input tokens.
Specifically, each patch is flattened to a vector of length $H_{t} \times W_{t}$ (we set $H_{t} = W_{t} =4$ in this paper), in which each element has a $C$ dimensional embedding.
This vector is then considered as a length-16 sequence of $C$ dimensional token, combined with a learnable position encoding.
To reduce the computational costs, we align the embedding dimension of each token with the dimension of the local representation path (we set $C=64$ in this paper), which is not identical to the current vision Transformers (For example, the embedding dimension $C$ of IPT \cite{IPT} is set to 576).
Then, we use these tokens as input to Transformers and output the global feature $F_{global}$.
To take the $n^{th}$ Transformer block for example, the input feature $F_{g}^{n-1}$ is multiplied by three learnable matrices to generate query $Q$, key $K$, and value $V$ matrices respectively.
\begin{equation}
\left\{ \begin{array}{l}
{Q = W_{Q} \times F_{g}^{n-1}} \\
{K = W_{K} \times F_{g}^{n-1}} \\
{V = W_{V} \times F_{g}^{n-1}} \\
\end{array}, \right.
\end{equation}
where $W_{Q}$, $W_{K}$, and $W_{V}$ denote three learnable matrices.
By calculating the correlation between $Q$ and $K$, we obtain global attention weights to aggregate information from different locations of $F_{g}^{n-1}$, thereby capturing the global information.
The global attention weights are subsequently multiplied with $V$ to obtain the weighted integrated features $F_{weighted}$.
\begin{equation}
F_{weighted} = softmax( \frac{QK^{T}}{\sqrt{d}})V,
\end{equation}
where $d = \frac{C}{N_{head}}$, $N_{head}$ denotes the number of attention head.
Finally, the $F_{weighted}$ is passed through a MLP to output the feature of the $n^{th}$ Transformer block. The output feature of $n^{th}$ Transformer block is formulated as:
\begin{equation}
F_{g}^{n} = H_{mlp}(F_{weighted}+F_{g}^{n-1}) + F_{weighted}+F_{g}^{n-1},
\end{equation}
where $H_{mlp}(.)$ denotes a MLP as feed forward network, which contains two fully connected layers.

\subsection{Self-calibrated Multi-path Fusion Network for Local Representation}
The local representation path in Figure \ref{fig:TANet} shows the architecture of the proposed SMFN, which contains $G$ self-calibrated multi-path fusion modules (SMFMs).
Each SMFM contains $I$ residual blocks (RBs), a feature calibration path, and the multi-path feature fusion part, which is shown in detail in Figure \ref{fig:SMFM}.
To take the $g^{th}$ SMFM for example, the input feature $F_{l}^{g-1}$ is first fed into a path consisting of $I$ RBs to extract deep feature $F_{res}$.
\begin{equation}
\begin{split}
F_{res} = H_{RB}^{g,I}(H_{RB}^{g,I - 1}( \cdots H_{RB}^{g,1}( F_{l}^{g-1} ) \cdots )),
\end{split}
\end{equation}
where $H_{RB}^{g,I}(.)$ denotes the function of $I^{th}$ RB of the $g^{th}$ SMFM.
Meanwhile, inspired by \cite{MCN}, another path produces the calibration weight $weight_{cal}$ through two 1$\times$1 convolution layers and a $Sigmiod$ function.
\begin{equation}
weight_{cal} = Sigmiod(H_{Conv}^{1\times1}(H_{Conv}^{1\times1}(F_{l}^{g-1})))
\end{equation}
The $weight_{cal}$ is then multiplied by $F_{l}^{g-1}$ to obtain the calibrated feature $F_{cal}$.
\begin{equation}
F_{cal} = H_{Mul}(weight_{cal}, F_{l}^{g-1}),
\end{equation}
where $H_{Mul}(.)$ denotes an element-wise product operation.
After obtaining $F_{res}$ and $F_{cal}$, we concatenate the two features along the channel axis and leverage a convolutional layer with a kernel size of 1$\times$1 to learn the fused feature representation. The fused feature $F_{fused}$ is formulated as:
\begin{equation}
F_{fused} = H_{Conv}^{1\times1}(H_{Cat}(F_{res}, F_{cal})).
\end{equation}
Finally, we fuse $F_{res}$, $F_{fused}$, and $F_{l}^{g-1}$ to produce the output feature $F_{l}^{g}$.
\begin{equation}
F_{l}^{g}\!= \!H_{Sum}(F_{res},\! F_{fused},\! F_{l}^{g-1}\!)\!=\!H_{SMFM}^{g}(F_{l}^{g-1}\!),
\end{equation}
where $H_{SMFM}(.)$ denotes the functions of $g^{th}$ SMFM.

In summary, the whole local representation path (i.e. SMFN) is formulated as
\begin{equation}
\begin{split}
F_{local} \!= \! H_{SMFM}^{G}( H_{SMFM}^{G - 1}(  \cdots  H_{SMFM}^{1}( F_{shallow} ) \\\cdots  )) = H_{local}(F_{shallow}) = H_{SMFN}(F_{shallow}),
\end{split}
\end{equation}
where $H_{SMFN}(.)$ denotes the functions of SMFN.

\section{Experiments and Discussion}

\subsection{Datasets and Implementation Details}

In this paper, we conduct extensive experiments on two available face datasets, namely FFHQ \cite{FFHQ} and CelebA \cite{CelebA}.
For FFHQ dataset, which is a high-quality face dataset with considerable variation in terms of facial attributes such as age and ethnicity as well as image background.
We randomly select 3800 images as the training dataset, 100 images as the validation dataset, 100 images as the testing dataset, and all HR images are resized to 256$\times$256 pixels as the ground truth.
For CelebA dataset, as a large-scale face attributes dataset, it consists of subjects with great diversities including large pose variations and background clutter.
Similar to the FFHQ dataset, we randomly choose 4000 images from CelebA dataset, and then use 3800 images for training, 100 images for validation and the rest of 100 images for testing. All HR images are resized to 216$\times$176 pixels.
Note that, we are committed to dealing with FSR tasks with large scaling factors (e.g., $\times$4 and $\times$8), which are the major challenges of the face image reconstruction.

Data augmentation is performed on the all training samples, which are randomly rotated by 90\textdegree, 180\textdegree, 270\textdegree, and flipped horizontally.
We initialize the learning rate as $2\times10^{-4}$ with reduced rate of 0.5 after every 60 epochs for two datasets.
Our model is trained by Adam optimizor with $\beta_{1} = 0.9$, $\beta_{2} = 0.999$, and $\epsilon = 10^{-8}$. We use Pytorch to implement our models with 2 RTX 2080 Ti GPUs.

\subsection{Evaluation Metrics}
The reconstruction results are evaluated with four evaluation metrics: Peak Signal to Noise Ratio (PSNR), Structural Similarity (SSIM) \cite{SSIM}, Learned Perceptual Image Patch Similarity (LPIPS) \cite{LPIPS}, and Mean Perceptual Score (MPS) \cite{AIM2020}.
PSNR and SSIM are standard evaluation metrics, which are widely used in low-level visual tasks.
LPIPS and MPS are the novel perceptual metrics to measure the perceptive quality of image, where the smaller value of LPIPS means more perceptual similarity.
MPS is the average of the SSIM and LPIPS, which is formulated as:
\begin{equation}
MPS = 0.5\times(SSIM+(1-LPIPS)).
\end{equation}

\subsection{Ablation Experiments}

\subsubsection{Verifying the effect of the local and global representation paths}
To demonstrate the effect of two paths in TANet, we set up three experiments, where the first experiment is to evaluate the performance of 4$\times$ and 8$\times$ SR on FFHQ dataset after removing local representation path in TANet, the second experiment is to evaluate the performance after removing global representation path in TANet, and the third experiment is to evaluate the performance of the original TANet.
Table \ref{tab:Verifying the effect of the local and global representation paths} shows the results of above three experiments, we can draw two conclusions: the first point is that the reconstruction performance is severely degraded when the local representation path is removed, because the limited local representation capability can not effectively predict the fine facial texture details; the second point is that the Transformer-based global representation path can provide considerable performance improvement by capturing the global facial structure information to constrain the local texture generation.
Note that, the number of SMFMs and RBs in the local representation path are both set to 10, which refers to \cite{SISN}, and the ablation study of Transformer blocks in the global representation path is elaborated in section 1 of the Technical Appendix.

%\begin{figure}[h]
%	\centering{\includegraphics[width=\linewidth]{image/ablation}}
%	\caption{Comparison with different number of Transformer blocks for 4$\times$ SR on FFHQ testing dataset.}
%	\label{fig:Ablation}
%\end{figure}
%
%
%\subsubsection{Exploring the optimal number of Transformer blocks in global representation path}
%\label{Exploring Transformer blocks}
%To explore the optimal number of Transformer blocks in global representation path, we design a set of experiments with different number of Transformer blocks for 4$\times$ SR on FFHQ dataset.
%Figure \ref{fig:Ablation} shows the experimental results of TANet with the number of Transformer blocks ranging from 5 to 12, we can find that when the number of Transformer blocks increases from 5 to 12, the performance of face reconstruction is optimal when the number of Transformer blocks is equal to 9, and the performance does not increase when the number of Transformer blocks increases to 10, 11, and 12.

\begin{table}[t]
	\begin{center}
		\caption{Verifying the effect of the local and global representation paths. "TANet w/o local" indicates that the local representation path is removed. "TANet w/o global" indicates that the global representation path is removed.}\label{tab:Verifying the effect of the local and global representation paths}
		\begin{tabular}{c|cccc}
			\hline
			\multirow{2}{*}{Methods}   &  \multicolumn{2}{c}{Scale = 4} &   \multicolumn{2}{c}{Scale = 8}     \\
			\cline{2-5}
			& PSNR & SSIM & PSNR & SSIM\\
			\cline{2-5}
			\hline
			\hline
			TANet w/o local & 30.16 & 0.8470 & 26.38  &  0.7271 \\
			TANet w/o global & 32.52 & 0.8895 & 28.50 &  0.7944  \\
			TANet & \textbf{32.61} & \textbf{0.8910} & \textbf{28.62} & \textbf{0.7972}  \\
			\hline
		\end{tabular}
	\end{center}
\end{table}

\begin{figure*}[h]
	\centering{\includegraphics[width=\linewidth]{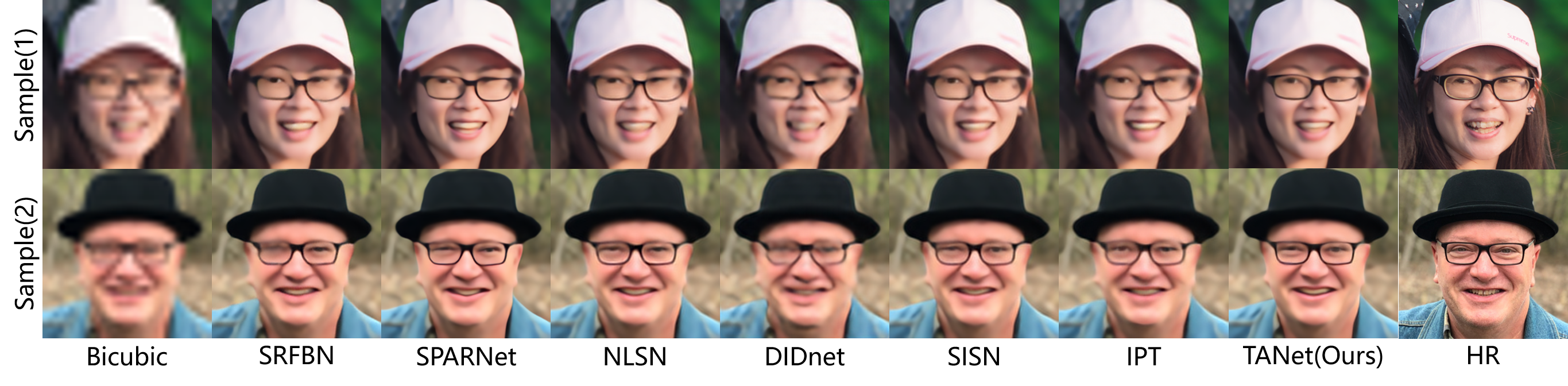}}
	\caption{Visual comparison for 8$\times$ SR with different methods. Two images from FFHQ testing dataset are selected as the test samples to show the reconstruction results of face images.}
	\label{fig:test sample_FFHQ}
\end{figure*}

\begin{table}[t]
	\begin{center}
		\caption{Comparison for 4$\times$ and 8$\times$ SR with the state-of-the-arts on FFHQ dataset. The bold denotes the best results. }\label{tab:ffhq256}
		\scalebox{0.9}{
			\begin{tabular}{c|c|cccc}
				\hline
				Methods & Scales  &  PSNR & SSIM & LPIPS & MPS   \\
				\cline{3-6}
				\hline
				\hline
				Bicubic & \multirow{8}{*}{$\times4$} & 29.60 & 0.8340 &0.3138 &  0.7601  \\
				SRFBN && 32.29 & 0.8859 & 0.1675 & 0.8592  \\
				SPARNet  & & 32.39 & 0.8887 & 0.1628 &  0.8630 \\
				NLSN & & 32.45 & 0.8885 & 0.1642 & 0.8622  \\
				DIDnet & & 31.95 & 0.8804  & 0.1729 & 0.8538 \\
				SISN & & 32.44 & 0.8882 & 0.1660 & 0.8611 \\
				IPT & & 32.15 & 0.8838 & 0.1722 & 0.8558 \\
				TANet(Ours) & & \textbf{32.61} & \textbf{0.8910} & \textbf{0.1590} & \textbf{0.8660} \\
				\hline
				\hline
				Bicubic & \multirow{8}{*}{$\times8$} & 25.88 & 0.7197 & 0.5102 &  0.6048  \\
				SRFBN & & 28.33 & 0.7890 & 0.2958 & 0.7466  \\
				SPARNet  & & 28.40 & 0.7937 & 0.2911 & 0.7513  \\
				NLSN && 28.37 & 0.7901 & 0.3086 & 0.7408   \\
				DIDnet & & 27.71 & 0.7705 & 0.3438 & 0.7134 \\
				SISN & & 28.40 & 0.7906 & 0.3048 & 0.7429 \\
				IPT & & 28.18 & 0.7841 & 0.3145 & 0.7348 \\
				TANet(Ours) & & \textbf{28.62} & \textbf{0.7972} & \textbf{0.2844} & \textbf{0.7564} \\
				\hline
		\end{tabular}}
	\end{center}
\end{table}

\begin{table}[t]
	\begin{center}
		\caption{Comparison for 4$\times$ and 8$\times$ SR with the state-of-the-arts on CelebA dataset. The bold denotes the best results. }\label{tab:celeba}
		\scalebox{0.9}{
			\begin{tabular}{c|c|cccc}
				\hline
				Methods & Scales  &  PSNR & SSIM & LPIPS & MPS   \\
				\cline{3-6}
				\hline
				\hline
				Bicubic & \multirow{8}{*}{$\times4$} & 28.76 & 0.8341 & 0.2729 & 0.7806  \\
				SRFBN & & 31.98 & 0.8945 & 0.1211 & 0.8867 \\
				SPARNet  & & 31.85 & 0.8930 & 0.1222 & 0.8854 \\
				NLSN & & 32.12 & 0.8971 & 0.1185 & 0.8893 \\
				DIDnet & & 31.55 & 0.8900 & 0.1333 & 0.8784 \\
				SISN & & 32.13 & 0.8968 & 0.1158 & 0.8905 \\
				IPT & & 31.89 & 0.8935 & 0.1227 & 0.8854 \\
				TANet(Ours) && \textbf{32.28} & \textbf{0.8992} & \textbf{0.1133} & \textbf{0.8930} \\
				\hline
				\hline
				Bicubic & \multirow{8}{*}{$\times8$} & 24.71 & 0.7004  & 0.4706 &  0.6149  \\
				SRFBN & & 27.51 & 0.7903 & 0.2409 & 0.7747 \\
				SPARNet & & 27.24 & 0.7853 & 0.2361 & 0.7746 \\
				NLSN && 27.59 & 0.7916 & 0.2502 & 0.7707 \\
				DIDnet && 26.81 & 0.7671 & 0.2838 & 0.7417 \\
				SISN && 27.58 & 0.7924 & 0.2437 & 0.7744 \\
				IPT && 27.51 & 0.7894 & 0.2495 & 0.7700 \\
				TANet(Ours) && \textbf{27.87} & \textbf{0.7999} & \textbf{0.2271} & \textbf{0.7864} \\
				\hline
		\end{tabular}}
	\end{center}
\end{table}

\subsection{Comparison with State-of-the-Arts}
To further verify the practicability, we compare our proposed method with currently state-of-the-arts, including the novel FSR methods SPARNet \cite{SPARNet}, DIDnet \cite{DIDnet}, SISN \cite{SISN}, the widely used algorithms in generic image SRFBN \cite{SRFBN}, NLSN \cite{NLSN}, and the pioneer of Transformer-based image processing (e.g., denoising, super-resolution and deraining) method IPT \cite{IPT}.

\subsubsection{Comparing on FFHQ dataset}
Table \ref{tab:ffhq256} lists the quantitative experimental results of different state-of-the-arts and proposed method on FFHQ testing dataset for 4$\times$ and 8$\times$ SR, it is clearly that our method outperforms these state-of-the-arts in four metrics.
In these quantitative results, we can find that the Transformer-based approach IPT does not achieve satisfactory results. The main reason for this result is that the Transformer-dominated architecture lacks local representation capability leading to texture information loss, which is evidenced by qualitative experiments.

Figure \ref{fig:test sample_FFHQ} shows the qualitative results of different methods for 8$\times$ SR on FFHQ testing dataset, we can discover that the results reconstructed by IPT are indeed blurrier than those of the method with high quantitative results.
However, the reconstructed face images from IPT can maintain the consistency of facial structure better.
For example, in Sample (1), the results reconstructed by SRFBN, SPARNet, and NLSN cause severe distortion in the eye region, compared with the result reconstructed by IPT, which maintains better structural information, thus strongly demonstrating the superiority of the Transformers in capturing the global facial structure.
Compared to all state-of-the-arts, our well-designed method can maintain the consistency of facial structure and the fidelity of local facial texture details restoration simultaneously by aggregating the global representation of Transformers and the local representation of CNNs.
More visual results are shown in section 3 of the Technical Appendix.

\begin{figure*}[h]
	\centering{\includegraphics[width=17cm]{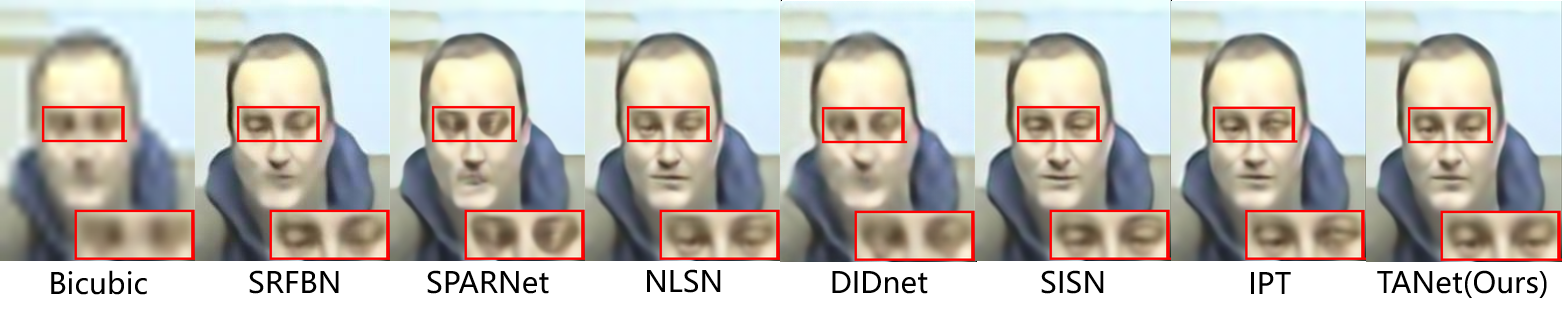}}
	\caption{Visual reconstruction results on real-world surveillance scenarios for 8$\times$ SR. Intuitively, our method reconstructs a clearer face and maintains more facial structure information.}
	\label{fig:realworld}
\end{figure*}

\subsubsection{Comparing on CelebA dataset}
Different from the FFHQ dataset, the original images in CelebA dataset are vague and contain some noise, which are more similar to the real-world scenarios.
Therefore, it is more challenging to conduct experiments on this dataset.
Table \ref{tab:celeba} summarizes the quantitative results of different methods on CelebA dataset, it is no surprise that the our TANet has achieved an overwhelming victory in all evaluation metrics, which prove that the proposed method has high generalization ability in various datasets.
Due to the space limitation, the visual reconstruction results for CelebA dataset are provided in section 3 of the Technical Appendix.

\subsection{Face Reconstruction and Recognition on Real-world Surveillance Scenarios}
In real-world surveillance scenarios, the resolution and quality of the face image captured by the imaging sensor on the surveillance camera are often low due to various complex and changeable situations (the imaging sensors are too far away from the face or the imaging quality of the sensor is poor).
Therefore, it is very important to utilize FSR to improve the quality of face images, thereby further promoting the performance of downstream tasks such as face detection and recognition.
In this section, we conduct extensive face reconstruction and recognition experiments in real-world surveillance scenarios dataset named SCface \cite{SCface} to compare the performance of different methods.

\subsubsection{Face Reconstruction}
To verify the effectiveness of our method in real-world surveillance scenarios, we select a very low-quality face image (image size is 21$\times$28 pixels) from SCFace dataset to evaluate the subjective reconstruction performance of different methods for 8$\times$ SR.
Visual comparison of reconstruction performance is shown in Fig. \ref{fig:realworld}, we can intuitively conclude that our proposed TANet can produce the super-resolved face image with more credible and clearer facial details.
More visual results are shown in section 3 of the Technical Appendix.

%\begin{figure}[t]
%	\centering{\includegraphics[width=4.8cm]{image/sample_facerec}}
%	\caption{A pair of samples for face recognition,  where the image on the left denotes a HD frontal face image captured by a SLR camera as a face feature sample, and the image on the right denotes a LR face image captured from a surveillance camera as a target image of the face recognizer.}
%	\label{fig:sample_facerec}
%	%\vspace{-0.4cm}
%\end{figure}

\begin{table}[t]
	\begin{center}
		\caption{Comparison results for matching accuracy and average similarity of face images reconstructed by different methods. The bold denotes the best result and the "-" means that the faces of all candidates are not detected.}\label{tab:facerec}
		\scalebox{0.9}{
		\begin{tabular}{c|c|c}
			\hline
			Methods & Matching Accuracy & Average Similarity \\
			\hline
			\hline
			LR image & - & -  \\
			Bicubic  & 3.0769\% &  0.186256   \\
			SRFBN  & 13.8462\% &  0.280387 \\
			SPARNet & 6.1538\% & 0.247244 \\
			NLSN & 11.5385\% & 0.272104 \\
			DIDnet & 8.4615\% & 0.237116 \\
			SISN &  9.2308\% & 0.264843   \\
			IPT &  15.3846\%  & 0.283265\\
			TANet (Ours) & \textbf{16.9231\%}& \textbf{0.292729}\\
			\hline
		\end{tabular}}
	\end{center}
\end{table}

\subsubsection{Face Recognition}
There are two ways to evaluate the performance of face reconstruction in real-world scenarios, the first one is the subjective reconstruction results shown in the previous section, and the second one is to compare the impact of the reconstructed faces on downstream tasks.
Inspired by \cite{SISN}, we use MTCNN \cite{MTCNN} as the face detector and MobileFacenets \cite{mobilefacenets} as the face recognizer to compare the impact of reconstruction results of different methods on face recognition.
Specifically, we use the high-definition (HD) frontal face images of the test candidates as the source face feature samples and the LR face image captured by the surveillance camera corresponding to the test candidates as the target images of the face recognizer.
%Figure \ref{fig:sample_facerec} shows a pair of samples for face recognition, where the HD image on the left represents the source face feature sample of this candidate and the image on the right represents the target image of the face recognizer.
The sample pair of visualization for face recognition and detailed process of face recognition are reported in section 2 of Technical Appendix.

However, the face recognition experiment implemented in \cite{SISN} suffers from the following shortcomings: (1) The sample size is too limited, using only six subjects as test candidates  (a total of 130 subjects); (2) The challenge is insufficient, using only the results of 4$\times$SR for the recognition task.
Therefore, to make the experiment more convincing, we use all 130 objects in the SCface dataset as test candidates and compare the results of different methods under 8$\times$SR.
Furthermore, we introduce the metric of matching accuracy, by setting the similarity threshold $o$ ($o=0.4$ in this paper) between each subject source face feature sample and the target recognized image to count the ratio of successful recognition, when the similarity is greater than or equal to $o$ is considered as successful recognition. Finally, the matching accuracy is obtained according to the total number of successful recognition.
Table \ref{tab:facerec} lists the face matching accuracy and average similarity of face images reconstructed by different methods for 130 test candidates after face recognition in real-world surveillance scenarios, we can draw two conclusions: (1) Transformers-based methods (i.e. IPT and TANet) can effectively improve the performance of face recognition by capturing the global facial structure; (2) The proposed TANet can further improve the performance of face recognition by efficiently aggregating the global representation of Transformers and the local representation of CNNs.

\section{Conclusion}
In this paper, we propose a novel paradigm for global FSR to fully explore and exploit the representation of global facial structure and local facial details features.
We design a TANet to maintain the consistency of global facial structure and the fidelity of local facial texture details restoration simultaneously by aggregating the global representation of Transformers and the local representation of CNNs.
However, our proposed method also has some limitations.
Although our method achieves the highest score for face recognition experiments in real-world surveillance scenarios, the matching accuracy of 16.92\% is still not feasible for practical applications.
%the main reason for this is that the quality of the reconstructed face images is still not high.
Therefore, improving the reconstruction performance in real-world scenarios will be the focus of recent research.

\section{Acknowledgments}
This work is supported by the National Natural Science Foundation of China (62072350, U1903214, 61971165, 62001334), Hubei Technology Innovation Project (2019AAA045), the Central Government Guides Local Science and Technology Development Special Projects (2018ZYYD059), 2020 Hubei Province High-value Intellectual Property Cultivation Project, the Wuhan Enterprise Technology Innovation Project (202001602011971), the Graduate Innovative Fund of Wuhan Institute of Technology (CX2020223).

% Use \bibliography{yourbibfile} instead or the References section will not appear in your paper
\bibliography{aaai22}

\end{document}